\documentclass{article}

\usepackage{PRIMEarxiv}

\usepackage[utf8]{inputenc} 
\usepackage[T1]{fontenc}    
\usepackage{hyperref}       
\usepackage{url}            
\usepackage{booktabs}       
\usepackage{amsfonts}       
\usepackage{nicefrac}       
\usepackage{microtype}      
\usepackage{lipsum}
\usepackage{fancyhdr}       
\usepackage{graphicx}
\usepackage{natbib}
\usepackage{arabtex}
\usepackage{utf8}
\setcode{utf8}
\graphicspath{{media/}}     

\title{Ensemble of Pre-trained Language Models and Data Augmentation for Hate Speech Detection from Arabic Tweets
\thanks{\textit{\underline{Citation}}: 
\textbf{Authors. Title. Pages.... DOI:000000/11111.}} 
}

\author{
  Kheir Eddine Daouadi, Yaakoub Boualleg, Kheir Eddine Haouaouchi \\
  Laboratory of Vision and Artificial Intelligence (LAVIA) \\
  Echahid Cheikh Larbi Tebessi University \\
  Tebessa, Algeria\\
  \texttt{\{kheireddine.daouadi, yaakoub.boualleg, Kheireddine.haouaouchi\}@univ-tebbesa.dz} \\
}

\begin{document}
\maketitle
Today, hate speech classification from Arabic tweets has drawn the attention of several researchers. Many systems and techniques have been developed to resolve this classification task. Nevertheless, two of the major challenges faced in this context are the limited performance and the problem of imbalanced data. In this study, we propose a novel approach that leverages ensemble learning and semi-supervised learning based on previously manually labeled. We conducted experiments on a benchmark dataset by classifying Arabic tweets into 5 distinct classes: non-hate, general hate, racial, religious, or sexism. Experimental results show that: (1) ensemble learning based on pre-trained language models outperforms existing related works; (2) Our proposed data augmentation improves the accuracy results of hate speech detection from Arabic tweets and outperforms existing related works. Our main contribution is the achievement of encouraging results in Arabic hate speech detection.

\keywords{Twitter \and Hate Speech Detection \and Arabic Tweet Classification \and Ensemble Learning \and Pre-traine language model \and Fine-tuning}

\section{Introduction}\label{sec1}
Nowadays, the use of social media has substantially increased in Arab countries, which has allowed more freedom for speech in different domains. Twitter for example is one of the leading social media where users can share short text of up to 280 characters optionally followed by a link, video, or photo, known as a tweet. This free micro-blogging allows users to subscribe, follow other users, share content, like other tweets, repost another tweet and reply to another tweet. Today, Twitter is booming, the service has more than hundreds of millions of users producing over five hundred million tweets per day \footnote{\url{https://www.omnicoreagency.com/twitter-statistics/}}. The population of Saudi Arabia and Oman are the most countries that used the platform. The Arab users generate 27.4 million tweets per day \cite{areej}. From that big number, we can assume that hate speech can spread easily and quickly through this platform.

The General Policy Recommendation no. 15 of the European Commission\footnote{\url{http://hudoc.ecri.coe.int/eng?i=REC-15-2016-015-ENG}} has described Hate Speech (HS) as ''the advocacy, promotion or incitement, in any form, of the denigration, hatred or vilification of a person or group of persons, as well as any harassment, insult, negative stereotyping, stigmatization or threat in respect of such a person or group of persons and the justification of all the preceding types of expression, on the ground of race, color, descent, national or ethnic origin, age, disability, language, religion or belief, sex, gender, gender identity, sexual orientation, and other personal characteristics or status''.

Today, through Twitter, researchers propose approaches for Arabic HS detection. Nevertheless, two major challenges faced in this context are the limited performance and the problem of imbalanced data, which make them suffer from severe over-fitting \cite{Founta2018}. Automatic HS detection from Arabic tweets using traditional machine learning classifiers like Naïve Bayes (NB), Support Vector Machine (SVM) and Random Forest (RF) have shown reasonable accuracy results. However, they are based on the handcrafted features calculated based on some pre-defined methods like Term Frequency (TF), Term Frequency-Inverse Document Frequency (TF-IDF), Bag of Word (BoW), etc. Recently, Long Short Term Memory (LSTM), Convolution Neural Network (CNN), and Bidirectional Encoder Representations from Transformers (BERT) have already shown good results for HS classification from Arabic tweets. 

In our study, we leveraged ensemble learning that relies on pre-trained language models for the Arabic language \cite{antoun2020arabert,abdul2020arabert} to perform HS classification. Specifically, we fine-tuned various Arabic BERT models trained on various Twitter data. In addition, we proposed a semi-supervised learning method based on previously labeled data. Our research contributions can be summarized as follows:
\begin{itemize}
    \item We leveraged various Arabic BERT models via transfer learning and fine-tuning to build our baseline classifier.
    \item We evaluated ensemble learning based on the leveraged Arabic BERT models.
    \item We propose a data augmentation method based on semi-supervised learning and previously labeled data.
\end{itemize}
The rest of this paper is organized as follows: Section ~\ref{RW} reviews the most recent related works on Arabic hate speech detection. In Section ~\ref{PR}, we provide details of our methodology. Section ~\ref{RD} presents and discusses the results of the conducted experiments. Conclusion and future works are presented in Section ~\ref{CF}.

\section{Related Works} \label{RW}
The emergence of the Twitter platform has encouraged a multitude of research avenues including topic detection \cite{da1,khairidp,dida}, organization detection \cite{dadifin,daouadi2018towards,lpkm2018organization}, and bot detection \cite{khairibotdf,jucs2020real}. Thanks to its importance, HS detection from Arabic tweets has drawn the attention of several researchers. In the literature, many systems have been developed to resolve this classification problem. They follow two main principal approaches: a traditional approach, and a deep learning approach.
\subsection{Traditional Approaches}
In this context, researchers have focused on manual features engineering based on local metadata and rely on the presence of tweet features like hashtags. Some examples of traditional approaches are briefly described in the following.

Besides, authors in \cite{Mubarak-a} focus on the problem of detecting offensive tweets. They use features from the Arabic word2vec model with SVM, this yielded an Accuracy result of 88.6\%.

Likewise, authors in \cite{husain2020a} classify tweets being offensive or not. The best experimental results are obtained using the trained ensemble learning in offensive language detection, and this yielded F1 score of 88\%, which exceeds the score obtained by the best single learner classifier by 6\%.

Furthermore, authors in \cite{husain2020c} examine the effect of the preprocessing step on offensive language and HS classification. They claimed that an intensive preprocessing technique demonstrates its significant impact on the classification rate. The optimal experimental results are obtained using BoW and SVM, yielding F1 score results of 89\% and 95\% for offensive language and HS classification.

In a different strategy, authors in \cite{mubarak-b} classify user accounts being abusive or not. The best experimental results are obtained using list-based methods (Sweed Words + Log Odds Ratio), and this yielded F1 score of 60\%.
\subsection{Deep Learning Approaches}
In this context, researchers follow two principal approaches are deep learning from scratch approaches and fine-tuning approaches. 
\subsubsection{Deep Learning from Scratch Approaches}
In this context, researchers have focused on automatic feature engineering. They used deep learning models such as CNN and LSTM. Some examples of deep learning approaches are briefly described in the following.

Authors in \cite{alharbia} classify Arabic tweets being offensive or not. They use features from n-gram and word embedding model. The best experimental results are obtained using LSTM. This yielded Accuracy result of 74\%.

In a similar, authors in \cite{farha2020} investigate the impact of using Multitask learning on offensive and HS detection. The showed that Multitask learning based on CNN-LSTM improve the accuracy results,  yielded macro F1-score of 90.4\% and 73.7\% for classifying offensive and HS tweets, respectively.

Likewise, authors in \cite{rachid2020} classify tweets being cyberbullying or not. They use features from the Arabic Word2vec model achieving. The best experimental result is obtained using a simple CNN-LSTM model, yielded F1-score of 84\%.

Furthermore, authors in \cite{husain2020b} use features based on TF-IDF to classify tweets from those of offensive or not. The best experimental results are obtained using a Bidirectional gated recurrent unit classifier, and this yielded a macro F1-score of 83\%.

In a similar, authors in \cite{abuzayed2020} classify tweets being HS or not based on the Arabic Word2vec model. They conducted a series of experiments using CNN, RNN and SVM. This yielded a macro F1-score of 73\%.
 
Besides, authors in \cite{faris2020} classify tweets from those of hateful and normal ones. They used CNN-LSTM with different versions of Arabic Word2vec, which yielded an F1-score of 71.68\%.
 
Likewise, authors in \cite{alsafari2020hate} investigate the effect of word embedding models and neural network architectures on the accuracy rate using (Offensive and Hate vs. Clean), (Hate vs. Offensive vs. Clean) and (Nationality hate vs. Clean vs. Religion Hate vs. Gender Hate vs. ethnicity hate vs. Offensive) classification tasks. The best results are achieved using CNN and word2vec based on the Skip-gram model, which yielded F1-score results of 70.80\% 75.16\% 87.22\%, and for six-class, three-class, and two-class classification tasks, respectively.
 
Besides, authors in \cite{alsafari2020-a} use the AraBERT embedding model with an ensemble of CNNs and Bidirectional LSTM (BiLSTMs) classifiers. The best results are achieved using the average-based ensemble approach, which yielded F1-score results of 91.12\% (CNNs), 84.01\% (CNNs), and 80.23\%(BiLSTMs), for two-class, three-class, and six-class classification tasks, respectively. 
 
In a similar, authors in \cite{alsafari2020hate} use CNN with the Multilingual BERT embedding model, which yielded F1 score results of 87.03\% 78.99\%, and 75.51\% for two-class, three-class, and six-class classification tasks, respectively.
 
Furthermore, authors in \cite{duwairi2021} conducted a series of experiments using CNN, BiLSTM-CNN, and CNN-LSTM. They reported three types of experiment: Binary classification (Hate or Normal), Ternary classification (Hate, Abusive, or Normal), and Multi-class classification (Misogyny, Racism, Religious Discrimination, Abusive, and Normal). In the binary classification task, the CNN model outperformed other models and achieved an accuracy of 81\%. In the ternary classification task, both the CNN, and BiLSTM-CNN models achieved the best accuracy of 74\%. While in multi-class classification task, the best results are achieved by CNN-LSTM, and BiLSTM-CNN, yielded an Accuracy of 73\%.
 
In similar, authors in \cite{alshaalan2020} classify tweets being hate or not, they compared four models: CNN, GRU, CNN-GRU, and BERT. Their experimental results show that CNN model gives the best performance with an F1-score of 79\%.
 
Similarly, authors in \cite{areej} classify tweets into five distinct classes: none, religious, racial, sexism and general hate. They used four deep learning models: LSTM, CNN-LSTM, GRU, and CNN-GRU. Their experimental results show that the hybrid CNN-LSTM model gives the best performance with an F1-score of 73\%.
 
Besides, authors in \cite{hassan2020} evaluate the combination of SVM, CNN, CNN-BiLSTM on offensive language and HS classification tasks. These models showed a significant performance with 90.51\% macro F1-score for offensive language detection and 80.63\% macro F1-score for hate speech detection.

In a different strategy, authors in \cite{alghanmi2020} use both Arabic BERT and Arabic Word2vec) embedding models classify tweets being normal, hateful, or abusive. The best accuracy result is achieved using CNN, which yielded an F1 score of 72.1\%.
 \subsubsection{Transfer Learning Approaches}
In this context, researchers have focused on transfer learning based on fine-tuning. The majority of transfer learning approaches focus mainly on the Arabic versions of BERT. Some examples of transfer learning approaches are briefly described in the following:
 
Authors in \cite{elmadany2020} evaluate Arabic BERT on offensive language and HS classification tasks. This yielded a macro F1-score of 82.31\% and 70.51\% for classifying HS and offensive language classification task.
 
Likewise, authors in \cite{abdellatif2020} evaluate the Universal Language Model Fine-tuning on the offensive language and HS classification task, this yielded macro F1-score of 77\% and 58\% for classifying offensive and HS tweets.
 
Besides, authors in \cite{aldjanabi2021} studied the impact multitask learning model built on top of a pre-trained Arabic BERT on offensive and HS classification tasks, which yielded F1-score of 92.34\% and 88.73\% for classifying offensive and HS tweets.
\subsection{Data Augmentation Methods} \label{DAM}
In this context, researchers have focused on generating synthetic data in order to face the challenge of imbalanced learning. Some examples of data augmentation methods are briefly described in the following:

Authors in \cite{husain2020c} investigate the impact of upsampling technique (duplicate tweets of the minority class to balance the class label) on offensive and HS classification tasks. They showed that upsampling decrease the result of F1-score results.

In similar, the authors in \cite{haddad2020} propose two data augmentation methods for offensive and HS classification. The first one uses an external augmenting technique by adding some offensive tweets from another labeled data. While the second one uses the random oversampling method by shuffling the words into HS tweets to create new samples. 

Likewise, authors in \cite{elmadany2020} investigate the impact of seed words and tweet emotion for automatic annotate tweets for offensive and HS classification tasks. They showed that the examined method has a positive and negative impact on the offensive and HS classification tasks, respectively.

In a different strategy, authors in \cite{alsafari2021} explore the role of semi-supervised built on unlabeled tweets for classifying tweets being offensive, hateful, and normal. They showed that an improvement of up to 7\% was achieved from using additional pseudo-labeled tweets.

\section{Proposed Approach} \label{PR}
\begin{figure}[h]
    \centering
    \includegraphics[width=10cm,height=15cm]{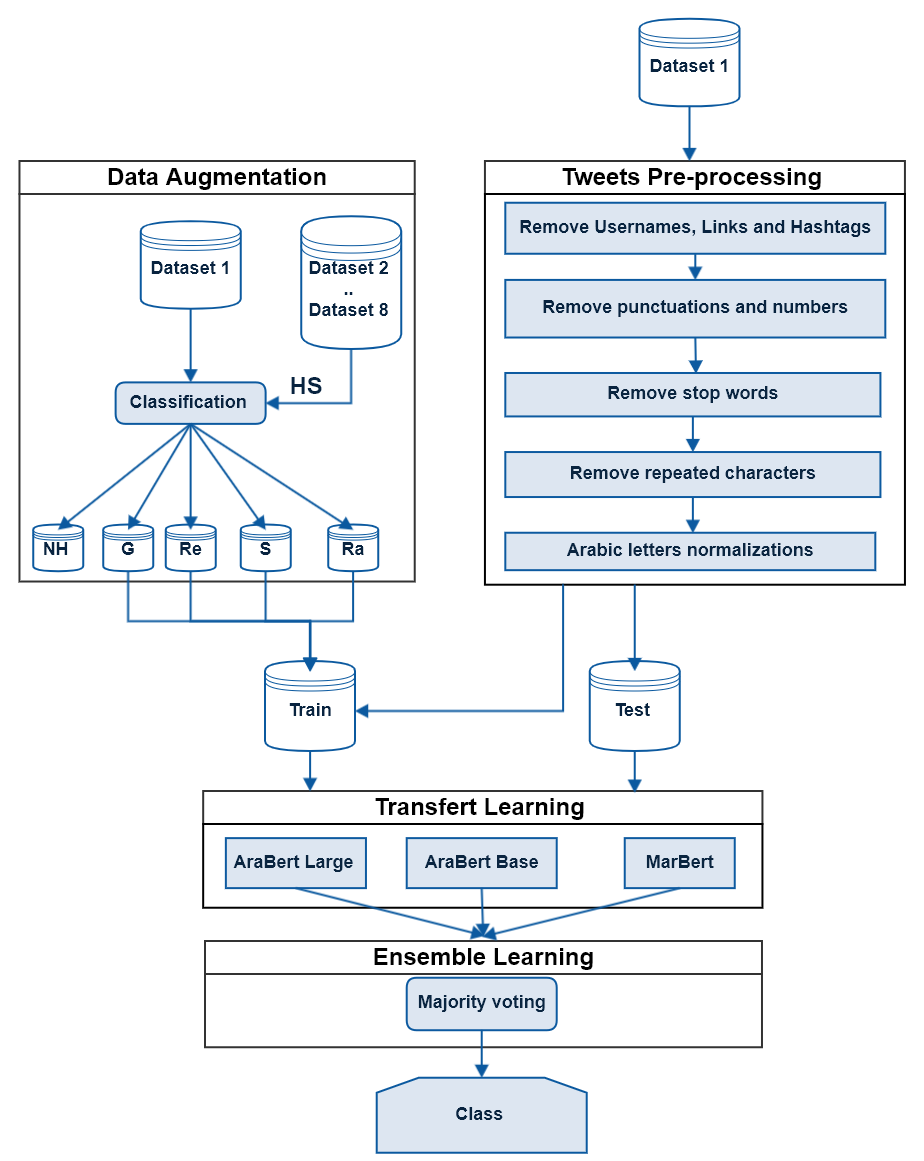}
    \caption{The process of transfer learning Corpus (NH: Normal, G: General hate speech, Re: Religious, S: Sexism, Ra: Racism).}
    \label{fig:model}
\end{figure}
To overcome the challenges previously mentioned in our related works, we propose a  novel approach for classifying hate  speech from Arabic tweets. As shown in Figure ~\ref{fig:model}, our proposed approach consists of four main steps are data augmentation, tweets pre-processing, transfer learning and ensemble learning.
\subsection{Data Augmentation}
The imbalance nature in data often has been disregarded in the major related works, which might have a substantial impact on the classification results. This problem has occurred in the HS detection where most labeled datasets are highly imbalanced. To face this issue, we propose semi-supervised learning built on previously manually labeled tweets. Firstly, we train the model using the state-of-the-arts datasets \cite{areej} labeled as (Non-hate, General HS, Sexism, Racial and Religious HS). The tweets labeled as religious HS in \cite{albadi2019investigating,alsafari2020hate} are added directly to the to the datasest of \cite{areej}. While the tweets previously labeled as HS in \cite{sun2019,mulki-etal-2019-l,Haddad19,alshaalan2020,alsafari2020hate,ousidhoum2019} are classified with the trained model, and the new label is used balance the first datasets.
\subsection{Tweets Preprocessing}
The inputs of our proposition are the textual content of tweets composed of raw tweet content. In this step, our main objective is to clean the textual content produce a more consistent and standard tweet. We performed some pre-processing tasks as follows: 
\begin{itemize}
    \item Removing the tweet features: user mentions ’@’, URLs, the word RT, \#, punctuation, special characters (emoticons), and numerical characters.
    \item Removing repeated characters, Arabic stop words, non-Arabic letters, new lines as well as diacritics.
    \item Arabic letters normalization: in the Arabic language, they are different variations for representing some letters which are:
    \begin{itemize}
        \item Letter \RL{ة} which can be mistaken and written as \RL{ه}, we normalized it to \RL{ه}.
        \item Letter \RL{أ} which has the forms \RL{ا}, \RL{آ}, \RL{إ}, \RL{أ} ), all these four letters are normalized into \RL{ا}.
        \item The Arabic dash that is used to expand the word \<مرحبا>) to \<مرحبـــــا>) has been removed.
        \item Letter \RL{ى} has been normalized to  \RL{ي}.
    \end{itemize}
\end{itemize}
All tweets are padded to the length of the longest tweet and will be used as an input for the following step.
\subsection{Transfer Learning}
The second step aims to use the pre-trained language models (AraBert-Large, AraBert-Base, and MarBert).

Traditional machine learning technology has matured to a point where it may be used in a variety of practical applications with considerable success. However, it have certain limits in some real-world circumstances. For example, gathering sufficient training data is prohibitively expensive, time-consuming, or impossible, necessitating the use of transfer learning methods. Transfer learning is a method of training artificial neural networks that depend on pre-trained models on specific tasks and data. Transfer learning assumes that if a pre-trained network solves one issue well, it may be utilized to tackle a similar but distinct problem with a little extra training. The reliance on a large amount of training data can be reduced in this way.

\begin{figure}
    \centering
    \includegraphics{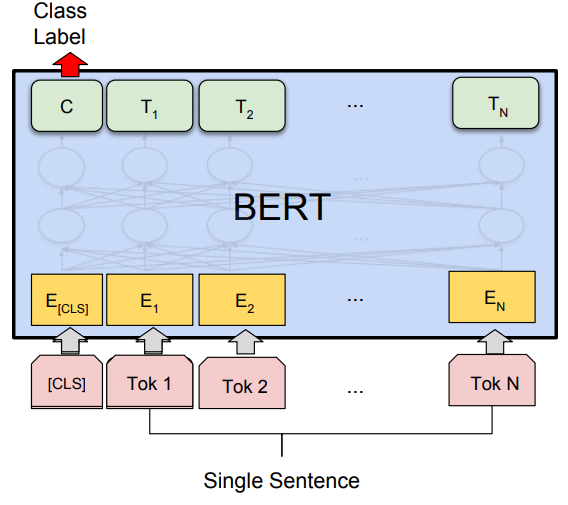}
    \caption{Single sentence classification using BERT.}
    \label{fig:cptr}
\end{figure}

Transfer learning is undeniably one of the most important aspects of language models. In the Natural Language Processing, the advent of Bidirectional Encoder Representation from Transformers (BERT) \cite{devlin2018bert} sparked a revolution. BERT is a deep learning model that has produced best-in-class performance on a wide range of natural language processing tasks.
BERT was taught using a technique known as Masked Language Modeling (MLM). The MLM operates by randomly concealing part of the existing tokens from the input. As a result, the masked word's original vocabulary id may be anticipated from the word's context. As a result, the MLM will be able to merge the left and right contexts. In addition to the MLM, BERT employs the Next Sentence Prediction task to train a competent language model that recognizes and understands sentence relationships. The BERT model architecture is a multi-layered bidirectional transformer encoder based on the \cite{devlin2018bert} original implementation. The transformers, it is suggested, are a collection of numerous nested layers (or blocks). An 'attention' layer is present in each layer/block. For each block, BERT uses twelve (12) different attention mechanisms, allowing tokens from the input sequence (e.g., sentences made up of word or sub-word tokens) to focus on the other token. In their paper, \cite{devlin2018bert} \cite{vaswani2017attention} offered the following two architectures: 
\begin{itemize}
    \item \textbf{BERT Base:} 12 layers (Transformer blocks), 12 attention heads, and 110 million parameters.
    \item \textbf{BERT Large:} 24 layers, 16 attention heads, and 340 million parameters.
\end{itemize}
A preprocessing phase, consisting of tokenization, should be conducted before feeding a raw sentence to BERT. The sub-word tokenization technique WordPiece employs a vocabulary initialization to cover all characters in the training data. WordPiece added the appropriate merge rules for tokenization later on. WordPiece learns positional embeddings with a sequence length of up to 512 tokens and creates embeddings with a 30,000 character vocabulary. English and BooksCorpus are used to train BERT models. The single sentence categorization using the BERT model is shown in Figure ~\ref{fig:cptr}. In Figure ~\ref{fig:cptr}, E stands for input embedding, Ti for contextual representation of token i (Tok I), and [CLS] for classification output, which summarizes all the hidden states outputted from all tokens in the input sentence.

\subsection{Ensemble Learning}
Ensemble learning is a broad machine learning meta-approach that tries to improve predictive performance by mixing predictions from several models. In our research, we compared two ensemble learning methods are majority voting and average voting described as follows.

Majority voting (also known as Hard Voting), each classifier votes for a class, and the class with the most votes win. The ensemble's anticipated target label is the mode of the distribution of individually predicted labels in statistical terms. This strategy is intentionally used to increase model performance, to out performs any one model in the ensemble.

Average voting (also known as Soft voting), where each classifier assigns a probability value to each data point that it belongs to. The predictions are totaled and weighted according to the relevance of the classifier. The vote is then cast for the target label with the largest sum of weighted probability.
\section{Experimental results and evaluation}\label{RD}
To fine-tune the transfer learning model, authors in \cite{sun2019} have recommended selecting from the values of the following parameters: learning rate, batch size, number of epochs described as follows:
\begin{itemize}
    \item The learning rate is a hyper-parameter that controls how much the model changes each time.
    \item The number of epochs is a hyper-parameter that specifies how many times the learning algorithm will iterate over the whole training dataset.
    \item The batch size is a hyper-parameter that specifies how many samples must be processed before the internal model parameters are updated.
\end{itemize}
The tenfold cross-validation approach was used to evaluate the performance of our proposed approach. The dataset was split into ten equal-sized segments while maintaining the balance of each class in the corresponding datasets. One of these parts was used as a testing and the remainder were used as training. This procedure was repeated 10 times and the averaging of the performance results was obtained across the ten repetitions of cross-validation. The performance measures used to evaluate our proposed approach are as described follows.
\begin{itemize}
    \item Recall (R), because it concentrates on the good examples, we should pay greater attention to that measure. For each class label, recall refers to the proportion of properly classified tweets to the number of tweets that belonged to that class but were mistakenly classified by the model. For example, the Recall of the Racial class is calculated as described in the following equation: 
       \begin{equation}
        R_{Racial}=\frac{correctly\_classified\_as\_Racial}{Total\_number\_of\_Racial\_tweets}
    \end{equation}
    \item Precision (P), this metric is likewise useful in our instance; it reflects the proportion of relevant retrieved tweets from a specific class. For example, the Precision of religious class is calculated as described in the following equation:
    \begin{equation}
        P_{Religious}=\frac{correctly\_classified\_as\_Religious}{Tweets\_classified\_as\_"Religious"}
    \end{equation}
    \item F1-Score is more useful than accuracy in our instance of unequal class distribution since it is the weighted average of precision and recall.
    \begin{equation}
        F1-Score=2 \times \frac{Precision \times Recal}{Precision + Recall}
    \end{equation}
    \begin{itemize}
        \item Micro : Calculate measures globally by counting the total true positives, false negatives, and false positives.
        \item Macro : Calculate measures for each label, and find their unweighted mean, this does not take class imbalance into account.
        \item Weighted : Calculate measures for each label and find their average weighted by support (the number of true instances for each label). This alters ‘macro’ to account for label imbalance.
    \end{itemize}
\end{itemize}
To validate our proposition, we used different Arabic HS datasets. In the following, a brief description of the corresponding datasets is presented.

We use the dataset annotated by \cite{areej} to evaluate our proposed classification approach. Table ~\ref{tab:1} shows an overview of the dataset we utilized in our research. The authors utilized the Especial library with a list of hashtags that activate and attract Twitter nasty content to collect the tweets. Because of this, balancing the amount of non-hate and hate tweets is unrealistic and does not reflect the true situation. To maintain a realistic and natural environment, the writers simply picked a list of hashtags that undoubtedly contains both non-hateful and hateful information. After that, two annotators manually annotated the obtained tweets to eliminate any annotator bias. A guide was supplied to the annotators to follow to separate the tweet classifications, resulting in a dataset of 11634 tagged tweets out of 37 k collected tweets. Table ~\ref{tab:1} shows that the tiny subset of tweets belongs to the racial hate speech category, whereas the large subset belongs to the non-hate category.
\begin{table*}[h]
    \centering
    \caption{Overview of the datasets used to evaluate our proposed classification approach (NH: Non-Hate speech, Se: Sexism hate speech, Re: Religious hate speech, GH: General hate speech, Ra: Racism hate speech, k denotes one thousand).}  \label{tab:1}
    \begin{tabular}{|l|c|c|c|c|c|c|c|c|c|}
    \hline
          & \textbf{NH} & \textbf{GH} & \textbf{Re} & \textbf{Ra} & \textbf{Se} & \textbf{Total} \\ \hline
        Number of tweets  & 8332 & 1397 & 722 & 526 & 657 & 11634\\ \hline
        Word count & 96.9 K & 17.2 K & 9.2 K & 6.6 K & 8.3 K & 138.3 K\\ \hline
        Unique words  & 29 K & 8.7 K & 4.9 K & 4 K & 4.4 K & 37.9 K\\ \hline
        Average words per tweet & 11.6 & 12.3 & 12.7 & 12.7 & 12.6 & 11.9\\ \hline
    \end{tabular}
\end{table*}

The statistics about the datasets used to evaluate our proposed method for data augmentation are presented in Table ~\ref{tab:2} and described as follows:
\begin{itemize}
    \item \textbf{OSACT} \cite{sun2019}: This dataset comprises 10K tweets that have been labeled for offensiveness (OFF or NOT OFF) and hate speech (HS or NOT HS).
    \item \textbf{L-HSAB} \cite{mulki-etal-2019-l}: L-HSAB is a collection of 5,846 Syrian and Lebanese political tweets categorized as normal, abusive, or hateful.
    \item \textbf{T-HSAB} \cite{haddad2020}: T-HSAB brings together 6,024 Tunisian comments that are classified as normal, abusive, or hateful.
    \item \textbf{GHDS} \cite{alshaalan2020}: A total of 9,316 tweets were classed as hateful, abusive, or normal in the dataset (not hateful or abusive).
    \item \textbf{RHS} \cite{albadi2019investigating}: These include 6000 Arabic tweets, 1000 for each of the six religious groups.
    \item \textbf{MCH} \cite{alsafari2020hate}: This dataset contains 1100 tweets that were gathered utilizing contentious accounts and hashtags.
    \item \textbf{MLMA} \cite{ ousidhoum2019}: This consists of multi-language and multi-aspect, the Arabic tweets include 3353 of hate and non-hate speech.
\end{itemize}
\begin{table}[h]
    \centering
    \caption{Overview of the datasets used to evaluate our proposed method for data augmentation.}
    \label{tab:2}
    \begin{tabular}{|c|c|}
    \hline
        \textbf{Datast} & \textbf{Hate tweets}\\ \hline
        MLMA & 460\\ \hline
        GHDS & 180\\ \hline
        RHS & 2759\\ \hline
        MCH & 1417\\ \hline
        OSACT & 491\\ \hline
        T-HSAB & 1078\\ \hline
        L-HSAB & 468\\ \hline
        All data & 6853\\ \hline
    \end{tabular}
\end{table}
\subsection{Fine-tuning }
The first set of our experiment is dedicated to determining the best hyper-parameter. The initial parameters used in our experiment are 2 epochs, 8 batch size, and 1e-5 learning rate.

The experiment presented in Table ~\ref{tab:3} shows the effect of the number of epochs. The F1-score metric is increased when the number of epochs is increased from 2 to 4 by 4.99\% in bert-base-arabertv02-twitter, but it is decreased when the number of epochs is further increased. Then, using bert-large-arabertv02-twitter, The F1-score metric is increased from 2 epochs, but it is decreased when the number of epochs is further increased. While when using MARBERT, the F1-score is increased by 0.16 when the number of epochs is increased from 2 epochs to 4 epochs, but it is decreased when the number of epochs is further increased
\begin{table*}[h]
    \centering
    \caption{The effect of the number of epochs on micro F1-score.}
    \label{tab:3}
    \begin{tabular}{|l|c|c|c|c|c|}
    \hline
        \textbf{Epochs}  & \textbf{2} & \textbf{3} & \textbf{4} & \textbf{5} & \textbf{10}  \\ \hline
        bert-base-arabertv02-twitter & 79.65 & 84.60 & \bf 84.64 & 84.28 & 84.14  \\ \hline
        bert-large-arabertv02-twitter & \bf 84.59 & 83.78 & 81.74 & 83.24 & 79.68  \\ \hline
        MARBERT & 83.98 & \bf 84.14 & 83.91 & 83.58 & 83.11  \\ \hline
    \end{tabular}
\end{table*}

The experiment presented in Table ~\ref{tab:4} shows the effect of the number of batch size. The F1-score is increased when the number of batch size is increased from 8 to 16 by 0.02 using bert-base-arabertv02-twitter, but it is decreased when the number of batch size is further increased. However, the F1-score metric is decreased when the number of batch size is further increased from 8 to 64 using bert-large-arabertv02-twitter and MARBET. 

\begin{table*}[h]
    \centering
    \caption{The effect of the number of Batch size on micro F1-score.}
    \label{tab:4}
    \begin{tabular}{|l|c|c|c|c|}
    \hline
        \textbf{Batch size} & \textbf{8} & \textbf{16} & \textbf{32} & \textbf{64}  \\ \hline
        bert-base-arabertv02-twitter & 84.64 & \bf 84.66 & 84.11 & 83.76  \\ \hline
        bert-large-arabertv02-twitter & \bf 84.59 & 84.08 & 83.66 & 81.88  \\ \hline
        MARBERT &  \bf 84.14 & 83.87 & 84.07 & 82.92  \\ \hline
    \end{tabular}
\end{table*}

The experiment presented in Table ~\ref{tab:5} shows the effect of the learning rate. The F1 score is decreased when the value of learning rate is increased from 1e-5  to 5e-5.

\begin{table*}[h]
    \centering
        \caption{The effect of the number of learning rate on micro F1-score.}
    \label{tab:5}
    \begin{tabular}{|l|c|c|c|c|c|c|c|c|}
    \hline
        \textbf{Learning rate} & \textbf{1e-5} & \textbf{2e-5} & \textbf{3e-5} & \textbf{4e-5} & \textbf{5e-5}\\ \hline
        bert-base-arabertv02-twitter & \bf 84.66 & 84.62 & 84.53 & 84.20 & 83.17\\ \hline
        bert-large-arabertv02-twitter & \bf 84.59 & 82.25 & 78.29 & 76.83 & 83.66\\ \hline
        MARBERT & \bf 84.14 & 83.42 & 81.83 & 80.55 & 79.68\\ \hline
    \end{tabular}
\end{table*}

In the summary, the optimal parameters of the corresponding models are shown in table ~\ref{tab:6}.

\begin{table}[h]
    \centering
    \caption{Optimal parameters of the corresponding pre-trained language model (E: Epochs, BS: Batch Size, LR: Learning Rate).}
    \label{tab:6}
    \begin{tabular}{|l|c|c|c|}
    \hline
        ~ & \textbf{E} & \textbf{BS} & \textbf{LR}  \\ \hline
        bert-base-arabertv02-twitter & 4 & 16 & 1e-5  \\ \hline
        bert-large-arabertv02-twitter & 2 & 8 & 1e-5  \\ \hline
        MARBERT & 3 & 8 & 1e-5  \\ \hline
    \end{tabular}
\end{table}
\subsection{The Effect of Ensemble Learning }
The second set of our experiments is dedicated to evaluating the ensemble learning methods (majority voting \& average voting). Table ~\ref{tab:7} present a comparison between single and ensemble models. It is important to note that ensemble learning models achieve the highest results, yielding weighted-average F1-score results of 85.48 and 85.10 using majority voting and averaged voting, respectively. 
In a summary, the Macro-Averaged, Micro-Averaged, and F1-score are reached 72.66\%, 85.47\% and 85.48\%, respectively.
\begin{table*}[h]
    \centering
    \caption{The effect of ensemble learning (Ma: Macro-Average, Mi: Micro-Average, W: Weighted-Average).}
    \label{tab:7}
    \begin{tabular}{|l|c|c|c|c|c|c|c|c|}
    \hline
        ~ & \textbf{NH} & \textbf{Se} & \textbf{Re} & \textbf{GH} & \textbf{Ra} & \textbf{Ma}   & \textbf{Mi} & \textbf{W}   \\ \hline
        bert-base-arabertv02-twitter & 92.42 & 70.26 & 79.89 & 60.72 & 50.48 & 70.76 & 84.66 & 84.69  \\ \hline
        bert-large-arabertv02-twitter & 92.47 & 67.60 & 79.74 & 60.82 & 47.81 & 69.69 & 84.59 & 84.46  \\ \hline
        MARBERT & 92.17 & 68.83 & 78.42 & 59.23 & 50.35 & 69.80 & 84.14 & 84.15  \\ \hline
        Average voting  & 92.60 & 70.86 & 81.53 & 61.21 & 52.37 & 71.71 & 85.08 & 85.10  \\ \hline
        Majority voting  & \bf 92.73 &  \bf 72.27 &  \bf 82.56 &  \bf 61.89 &  \bf 53.85 &  \bf 72.66 &  \bf 85.47 &  \bf 85.48  \\ \hline
    \end{tabular}
\end{table*}
\subsection{The Effect of Data Augmentation}
The third set of our experiments is delegated to evaluating our proposed method for data augmentation.
Firstly, the tweets labeled as religious hate speech by \cite{albadi2019investigating,alsafari2020hate} are directly added to the training data. Second, we used semi-supervised learning using the trained model with the datasets presented in Table ~\ref{tab:2}. Table ~\ref{tab:8} present the results of the classification task. 

\begin{table}[h]
    \centering
    \caption{Prediction results.}
    \label{tab:8}
    \begin{tabular}{|l|l|l|l|l|}
    \hline
        \textbf{Predicted class} & \textbf{GH} & \textbf{Re} & \textbf{Se} & \textbf{Ra}  \\ \hline
        Number tweet & 1034 & 950 & 863 & 926  \\ \hline
    \end{tabular}
\end{table}

The experiments presented in Table ~\ref{tab:9} shows a comparison of the proposed approach using data augmentation and without data augmentation. However, the F1-score result is increased from 85.48 to 85.65 when the training data is augmented through or proposed method.  

\begin{table*}[h]
    \centering
    \caption{The effect of our proposed data augmentation method.}
    \label{tab:9}
    \begin{tabular}{|l|c|c|c|c|c|c|c|c|}
    \hline
        ~ & \textbf{NH} & \textbf{Se} & \textbf{Re} & \textbf{GH} & \textbf{Ra} & \textbf{Ma} & \textbf{Mi} & \textbf{W} \\ \hline
        Without Data Augmentation & \bf 92.73 & 72.27 & 82.56 & 61.89 & 53.85 & 72.66 & 85.47 & 85.48  \\ \hline
        With Data Augmentation  & 92.41 & \bf 72.58 & \bf 83.50 & \bf 63.27 & \bf 57.25 & \bf 73.80 & \bf 85.60 & \bf 85.65  \\ \hline
    \end{tabular}
\end{table*}
\subsection{Comparison with Existing Data Augmentation Methods}
The fourth set of our experiments is dedicated to comparing our proposed methods with the relevant related works. Notably, from the experiment shown in Table ~\ref{tab:10}, the F1-score results of different classes is presented as follows:
\begin{itemize}
    \item \textbf{Normal:} The F1-score of the latest state-of-the-art data augmentation methods is ranging from [91.65\% – 92.36\%] while the F1-score of our proposed data augmentation methods yielded 92.41\%. 
    \item \textbf{General hate speech:} The achieved results from our proposed data augmentation method is increased by about 3.19\% when compared with the latest state-of-the-art.
    \item \textbf{Sexism:} The F1-score of the latest state-of-the-art data augmentation methods ranges from [66.38\% – 70.52\%], whereas our proposed data augmentation method achieved 72.58\%.
    \item \textbf{Racial:} Our proposed data augmentation method increases the F1-score from [44.15\% - 50.99\%] in comparison with the latest state-of-the-art, which yielded 57.25\%.
    \item \textbf{Religious hate speech:} The F1-score of the latest state-of-the-art data augmentation methods is ranging from [78.72\% - 79.89\%] while our proposed data augmentation method yielded F1-score of 83.50\%. 
\end{itemize}
In a summary, the Macro-Averaged, Micro-Averaged, and Micro-Averaged F1-scores are reached 73.80\%, 85.60\%, and 85.65\%, respectively.
\begin{table*}[h]
    \centering
        \caption{Comparison of our proposed data augmentation method with the latest state-of-the-art methods.}
    \label{tab:10}
    \begin{tabular}{|l|c|c|c|c|c|c|c|c|}
    \hline
        ~ & \textbf{NH} & \textbf{S} & \textbf{Re} & \textbf{G} & \textbf{Ra} & \textbf{Ma}   & \textbf{Mi} & \textbf{W} \\ \hline
        Without Data Augmentation & \bf 92.73 & 72.27 & 82.56 & 61.89 & 53.85 & 72.66 & 85.47 & 85.48  \\ \hline
        \cite{husain2020c} & 92.04 & 70.12 & 79.89 & 59.45 & 50.99 & 70.50 & 84.20 & 84.28  \\ \hline
        \cite{alsafari2021} & 92.21 & 70.52 & 79.42 & 60.08 & 50.60 & 70.57 & 84.29 & 84.45  \\ \hline
        \cite{liu2020}& 92.36 & 66.38 & 78.72 & 59.48 & 44.15 & 68.22 & 84.08 & 83.92  \\ \hline
        \cite{cao2020} & 91.65 & 67.50 & 79.20 & 57.01 & 49.07 & 68.89 & 83.52 & 83.43  \\ \hline
        Ours  & 92.41 & \bf 72.58 & \bf 83.50 & \bf 63.27 & \bf 57.25 & \bf 73.80 & \bf 85.60 & \bf 85.65  \\ \hline
    \end{tabular}
\end{table*}
\subsection{Comparison with Existing Classification Approaches}
The last set of our experiment is dedicated to comparing our proposed classification approach with the latest related works.

While comparing our proposed approach with the other baselines in Table ~\ref{tab:11}, the accuracy rate of our proposition is the highest. This is due to our use of ensemble learning and the the data augmentation method.
 When compared with CNN, LSTM, current DL hate speech classification models require to huge amount of labeled data to achieve good performance results.
 In contrast, does not require a huge amount of labeled data to achieve good performance results. 
When compared to SVM, NB, and Bagging, ours has the benefit of not requiring handcrafted features to be designed and extracted.
\begin{table*}[h]
    \centering
    \caption{Comparison between our proposed classification approach and the latest related works.}
    \label{tab:11}
    \begin{tabular}{|l|c|c|c|c|c|c|c|}
    \hline
        \bf Work & \bf NH & \bf Se & \bf Re & \bf GH & \bf Ra & \bf Ma & \bf W  \\ \hline
        \cite{mulki-etal-2019-l} & 0.85 & 0.21 & 0.10 & 0.12 & 0.09 & 0.27 & 0.64  \\ \hline
        \cite{husain2020a} & 0.86 & 0.42 & 0.50 & 0.25 & 0.24 & 0.45 & 0.71  \\ \hline
        \cite{chowdhury2020}  & 0.87 & 0.41 & 0.54 & 0.30 & 0.29 & 0.48 & 0.73  \\ \hline
        \cite{areej} & 0.86 & 0.43 & 0.59 & 0.31 & 0.25 & 0.49 & 0.73  \\ \hline
        \cite{alsafari2020hate} & 0.85 & 0.42 & 0.56 & 0.45 & 0.20 & 0.50 & 0.73  \\ \hline
        \cite{haddad2020} & 0.87 & 0.50 & 0.64 & 0.49 & 0.27 & 0.55 & 0.76  \\ \hline
        \cite{alghanmi2020} & 0.86 & 0.44 & 0.59 & 0.47 & 0.22 & 0.52 & 0.74  \\ \hline
        \cite{faris2020} & 0.87 & 0.37 & 0.47 & 0.28 & 0.26 & 0.45 & 0.72  \\ \hline
        \cite{alsafari2020-a} & 0.87 & 0.49 & 0.56 & 0.47 & 0.24 & 0.53 & 0.75  \\ \hline
        \cite{Mubarak-a} & 0.88 & 0.49 & 0.67 & 0.41 & 0.37 & 0.57 & 0.77  \\ \hline
        Ensemble Learning & \bf 0.93 & 0.72 & 0.83 & 0.62 & 0.54 & 0.73 & 0.85  \\ \hline
        Ensemble Learning + Data Agmentation  & 0.92 & \bf 0.73 & \bf 0.84 & \bf 0.63 & \bf 0.57 & \bf 0.74 & \bf 0.86  \\ \hline
    \end{tabular}
\end{table*}

\section{Conclusion}\label{CF}
Today, the increase in the spread of HS on social media is a global alarm. Efforts are now taking place to control and avoid HS. This research investigates the impact of ensemble learning that incorporates three different pre-trained Arabic language models with semi-supervised learning built on previously labeled datasets. The leveraged models are used to cover classical Arabic texts and their dialects. Moreover, using pre-trained language model will facilitate the process of learning content representations. Experiments results show that: (1) ensemble learning based on pre-trained language models outperforms existing related works; (2) Our proposed data augmentation improves the accuracy results of hate speech detection from Arabic tweets and outperforms existing related works.

As future works, we plan to follow numerous directions. First, we want to focus on the contextual embedding model and try to modify its vocabulary to support the HS detection task. Second, we plan to address the task of classifying HS in the Algerian dialect.

From a research perspective, we will leverage our proposed systems to classify and analyze Arabic Twitter discussions to determine the extent of HS conversations with public discourse and to understand how their sophistication and capabilities evolve over the time.

\end{document}